\documentclass[a4paper]{article}

\pdfoutput=1

\usepackage{INTERSPEECH2019}
\usepackage[dvipsnames]{xcolor}
\usepackage{listings}
\usepackage{algorithm}
\usepackage{algpseudocode}
\usepackage{amsmath}
\usepackage{url}

\DeclareMathOperator*{\argmax}{arg\,max}

\usepackage{array}
\usepackage{booktabs}
\title{Prosodic Phrase Alignment for Machine Dubbing} 
\name{Alp \"{O}ktem$^{1,2,\ast}$\thanks{$^*$Alp \"{O}ktem is currently at Translators without Borders.}, Mireia Farr\'us$^2$, Antonio Bonafonte$^{3,\ast\ast  }$\thanks{$\ast\ast$ Antonio Bonafonte is currently at Amazon Research, Cambridge, UK (bonafont@amazon.com)}
}

\address{
  $^1$Col$\cdot$lectivaT SCCL, Barcelona, Spain\\
  $^2$Universitat Pompeu Fabra, Barcelona, Spain\\
  $^3$Universitat Polit\`ecnica de Catalunya, Barcelona, Spain}
\email{\{alp.oktem, mireia.farrus\}@upf.edu \\
antonio.bonafonte@upc.edu}

\makeatletter
\def\blfootnote{\xdef\@thefnmark{}\@footnotetext}
\makeatother

\begin{document}

\maketitle

\begin{abstract}
Dubbing is a type of audiovisual translation where dialogues are translated and enacted so that they give the impression that the media is in the target language. It requires a careful alignment of dubbed recordings with the lip movements of performers in order to achieve visual coherence. In this paper, we deal with the specific problem of prosodic phrase synchronization within the framework of machine dubbing. Our methodology exploits the attention mechanism output in neural machine translation to find plausible phrasing for the translated dialogue lines and then uses them to condition their synthesis. Our initial work in this field records comparable speech rate ratio to professional dubbing translation, and improvement in terms of lip-syncing of long dialogue lines.
\end{abstract}


\section{Introduction}

%Audiovisual translation
Introduction of machine mediated methods to audiovisual translation domain has made it possible to obtain transcriptions and translations for multimedia without the huge manual labor that they used to demand. This is especially useful for online video streamers who publish often and lack access to professional translation services. By making their videos transcribed, one could augment visibility to their content; and by making them translated, they can further make them accessible to audiences of different languages. While subtitles remain as the most accessible method of audiovisual translation, it does not suffice as a medium for certain viewers with visual impairments or reading difficulties. Also, due to various cultural and historic reasons, subtitles are not the preferred form of audiovisual translation in certain linguistic communities \cite{RUPEREZMICOLA2019487}. 

%View increase citation: https://www.3playmedia.com/customers/case-studies/discovery-digital-networks/

The alternative to subtitling is \textit{dubbing}, where the dialogues in an audiovisual are translated in spoken form to replace its original audio track. It is a method widely used for the translation of e.g.~movies, commercials and video games. However, unlike subtitling, there are relatively few automated solutions for it. This can be explained mainly through the further complexities that the spoken language output introduces. In dubbing, the translated content does not only need to convey the linguistic and semantic information but also needs to deliver them within a similar prosodic context that matches the unchanged visual content. A key factor in dubbing, referred to as \textit{synchronization} or \textit{lip-sync}, deals with the reproduction of timing, phrasing and phonetic content of the original speech segments in the target language to match with the lip movements of the original performers~\cite{luyken}. This factor is especially relevant in the cases where the performer is on-screen and their articulations can be visually traced. In a majority of the cases, dubbing translation is made in such a way that it (1) contains a comparable number of syllables, (2) follows a similar phrasing and pausing structure, (3) matches mouth articulation movements (like opening, closing etc.) \cite{chaume2012audiovisual}. These requirements are especially important in movie-domain dubbing scenarios where the final content needs to sound and feel realistic in the target language. 

%There are many factors to think about for realizing machine dubbing. but we focus on...

In this work, we focus on the modelling of the cross-lingual prosodic phrasing structure in machine dubbing. In order to carry out experimentation, a prototype pipeline for obtaining dubbed versions of speaker segments of a TV show is presented. Our aim is not only to translate the spoken segments and synthesize them, but also to make the translated and synthesized segments to match the original durations and phrasings. The problem that requires a soft-alignment between translations is tackled by a simple mechanism that exploits the attention mechanism output during neural machine translation. Later, the durations of the input phrases are used to cue the text-to-speech (TTS) system so that the translated output is synthesized in a way that it respects the prosodic phrasing structure of the original take. 

\section{Automating Dubbing}
Automatic dubbing has been previously proposed in various works without any translation back-end \cite{czech2008, czech2010, subtts2009, Ljunglof:2012}. General motivation behind these works is to generate an additional track for a given audiovisual for aiding viewers with auditory or visual impairment. This approach uses the segmented transcriptions in already available subtitles of the audiovisual to generate a new audio track with synthesized speech. This additional track containing clearer speech is laid over the original track of the media, which is either muted or turned down. 

The only durational constraint dealt in previous works is the synchronisation of the synthesized segments with their respective subtitle timestamps. This is obtained by placing the synthesized segment at the beginning of each subtitle entry and then adjusting its speech rate only if it exceeds the duration of the subtitle entry. Matou\u{s}ek et al. performs an online and offline TTS time compression technique in order to achieve this \cite{czech2010}. Although this methodology can be useful in news or documentary domain, it would lead to limitations for obtaining realistically synthesized segments needed in e.g.~movie domain. For instance, subtitle segments that are considered as independent would lead to prosodically incoherent syntheses of long dialogue lines. Also, subtitle segmentation heeds prosodic phrasing only to some extent. Precise prosodic phrasing and phonetic alignment are not given attention in subtitle format translations. 

This paper, to the best of our knowledge, is the first study that approaches the automation of realistic cross-lingual dubbing. We outline a methodology for a possible application that could prove useful for media content creators and focus on the aspect of prosodic phrase synchronization, which is one of the core elements in dubbing.

\section{Synchronization in Movie-domain Dubbing}

As synchronization in dubbing is most relevant in movies, we base our analyses and experimentation on this domain. 

\subsection{Audio Data}
Both for analysis and experiments, we have used the \textit{Heroes Corpus} \cite{Oktem2018}, which is a collection of original and dubbed dialogue segments from a North American TV series. The series is originally in American English and is dubbed into many languages including Spanish. Heroes Corpus contains 7000 parallel English and Spanish single speaker audio segments, together with their transcripts and prosodic annotations. Each audio segment contains a dialogue line of an actor varying between 0.1 to 18 seconds (2.43s in average) uttering a minimum of 1 to a maximum of 45 words (5.43 in average). The prosodic annotations contain word alignment information with word-level silent pause duration information specified. 

% As subtitling and dubbing processes are normally carried out independently, it is usually the case that the dubbing script is not the same as the subtitle script. \"Oktem et al. \cite{Oktem2018} carried out a manual correction process to ensure that dubbed audio segments match with their transcripts. \alp{$\leftarrow$ paragraph to be taken out if space needs to be saved}

\subsection{Prosodic phrase alignment}

We have defined the prosodic phrases that are studied in this paper as voiced segments terminated with a silent pause. Although this is a partial definition of a prosodic phrase, it suffices for modelling of voiced segment alignment of dubbing phrases. 

%pause ref: http://www.ling.sinica.edu.tw/eip/FILES/publish/2007.12.20.44184511.458103.pdf
A corpus-level analysis has been performed to see to what extent pausing is reflected between the original and dubbed segments of the dataset. This was done by measuring if a paused interval overlaps in both original and dubbed segment timewise. It was seen that for any pause longer than 0.1 seconds, it was 70\% likely that it will be reflected in the dubbing translation. For pauses of 0.2 seconds in length this was 81\% and for pauses of 0.4 seconds it was 91\%, showing the attention given for the reflection of long pauses in dubbing movie dialogues. 

% \begin{figure}[t]
%   \centering
%   %\includegraphics[width=\linewidth]{img/}
%   \caption{Prosodic phrase alignment in a dubbed segment of Heroes Corpus.}
%   \label{fig:aligned_segs}
% \end{figure}

% \begin{figure}[t]
%   \centering
%   \includegraphics[width=\linewidth]{img/pause_reflection1.pdf}
%   \caption{Distribution of reflected pause intervals corresponding to minimum pause duration threshold}
%   \label{fig:pausecorr}
% \end{figure}

Another key issue in dubbing is making sure that the translated content fits the original take length-wise. This is done by making sure that the translation contains more or less a similar number of syllables with a careful word selection \cite{chaume2012audiovisual}. In our dataset, we have measured that the average ratio of the number of syllables in parallel segments is 1.31 with a standard deviation of 0.39, showing a higher speech rate for Spanish.
\section{Machine Dubbing Synchronization}

\begin{figure*}[t]
  \centering
  \includegraphics[width=\linewidth]{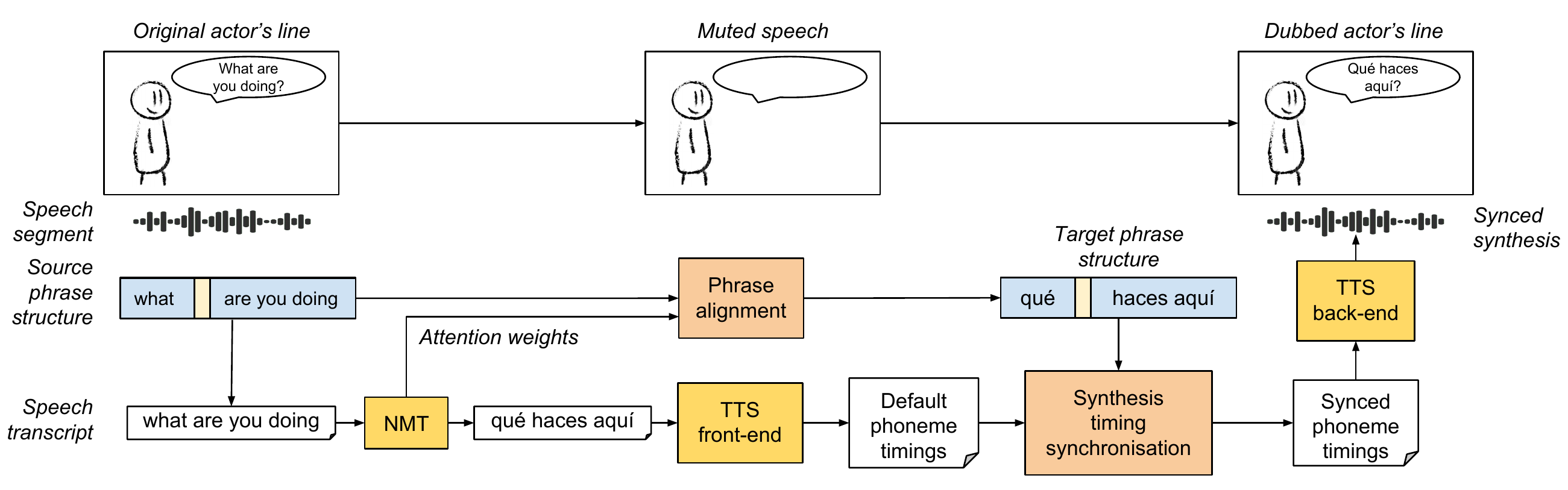}
  \caption{Prototype machine dubbing pipeline with prosodic phrase synchronisation.}
  \label{fig:pipeline}
\end{figure*}

The prototype machine dubbing setup that forms the basis of our synchronization methodologies is illustrated in Figure \ref{fig:pipeline}. The pipeline performs the translation of segmented and transcribed dialogue lines in the original version of the movie, like in the example below:

\begin{lstlisting}
MATT: What are you doing in here? Where are we?
MOLLY: This is where he keeps me.
MATT: What?
\end{lstlisting}

Having timestamp and word alignment information on these segments makes it possible to infer the prosodic phrasing structure in the dialogue line where silent pauses that fall between the uttered words are considered as prosodic phrase (PP) cut-off points. In order to differentiate between articulatory effects  and linguistically motivated pauses, we have determined a pause duration threshold of $250~ms$ following the conventions of \cite{deputy, goldman, oliviera}. 

% Once PP boundaries are determined, the words that fall into each phrase are labelled with a prosodic phrase id. To illustrate, words ($w_i$) in the input sentence $\langle w_1 \: w_2 \: SP \: w_3 \: w_4 \: SP \: w_5 \rangle$ fall into three prosodic phrase labels $\{ l_1, l_2, l_3 \}$ dictated by the silent pauses ($SP$). This leads to a PP label sequence of $\langle l_1, l_1, l_2, l_2, l_3 \rangle$.

Dialogue lines are translated using a neural machine translation (NMT) system. After that, translated phrases are synthesized using a text-to-speech (TTS) system with durational cues that reflect the source utterance. These durational and segmentation cues are determined by the two methodologies explained in the following two subsections.

\subsection{Cross-lingual phrase alignment using attention}
The first stage of phrasing synchronization deals with the determination of the PP boundaries in the target translation. The output of the machine translation is a sequence of tokens (words and punctuation) which needs to be split up into phrases that reflect the original input sentence structure. This necessitates a modelling of alignment between the input and output sentence. 

\begin{figure}[t]
  \centering
  \includegraphics[width=0.9\linewidth]{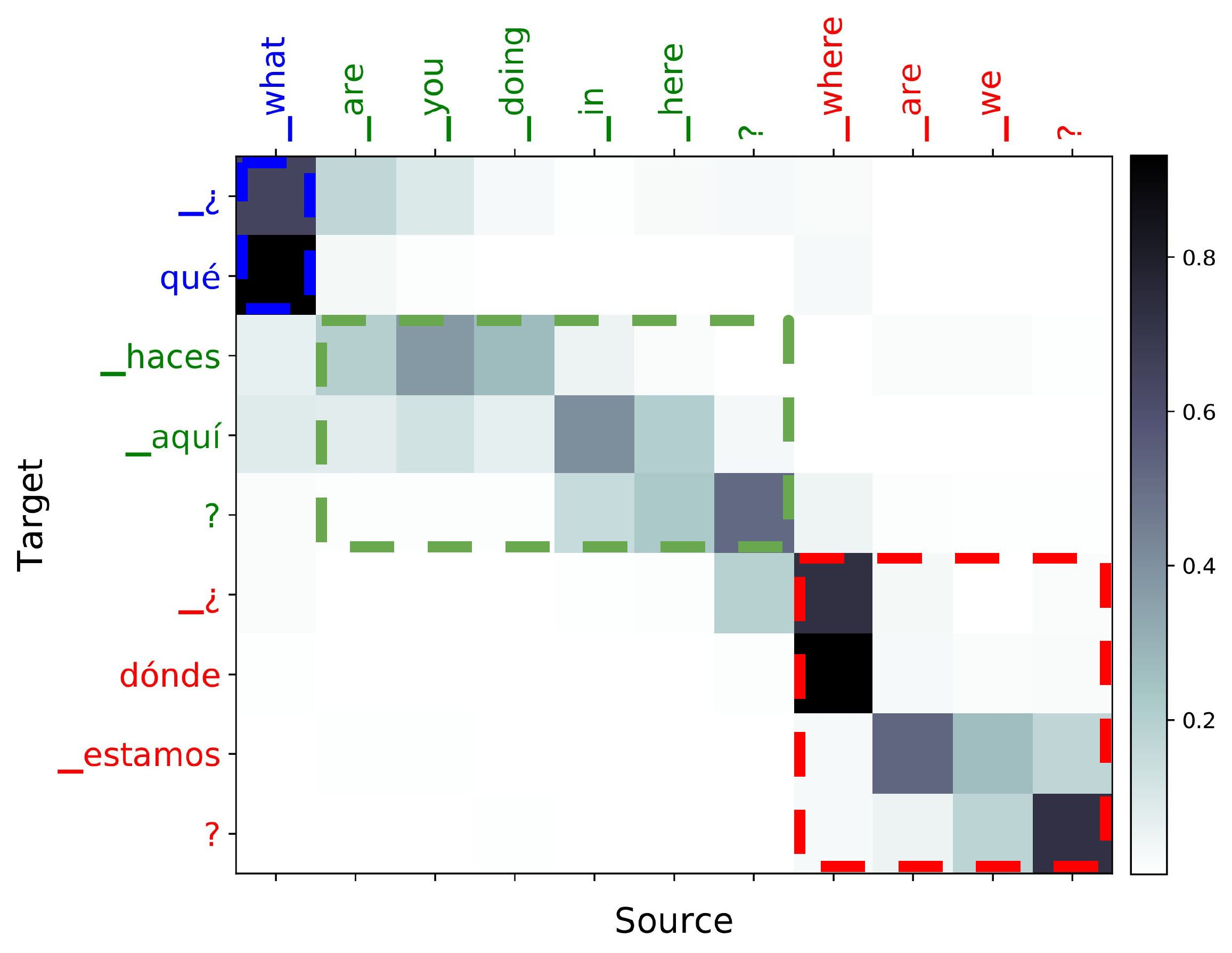}
  \caption{Visualization of attention output during translation of a sample from the dataset. Tokens belonging to different prosodic phrases are denoted with the colors. Attention weights are higher in the areas where source and target phrases align. }
  \label{fig:attention}
\end{figure}

Our methodology is based on exploiting the attention weights that is used in neural network based machine translation (NMT). Attention was first introduced for encoder-decoder based NMT \cite{sutskever} as a mechanism to help focus on different parts of the input at each step of decoding \cite{bahdanau}. It has also been shown that it represents a structure relating with the alignment between the input and output phrases \cite{hamidreza}. Designed to learn the relevance between parts of input and target sequence in training, it can be accessed during inference for obtaining alignment information between the source and predicted sentence. Figure \ref{fig:attention} shows the visualization of the attention weights output in translation. The alignment of the phrases can be traced on the distribution of attention weights. 

%In the case of PP alignment, it is not necessary to get precise alignment between words but only a partitioning of the target translation that matches the source phrasing. 

We define the cross-lingual phrase alignment process as follows: Given a source sentence with tokens $W_e:\langle w_e^1,w_e^2,\ldots,w_e^M \rangle$ and their respective PP labels $L_e:\langle l_e^1,l_e^2,\ldots,l_e^M \rangle$, find the best PP label sequence $L_f: \langle l_f^1,l_e^2,\ldots,l_e^N \rangle$ of the translated sequence of tokens $W_f: \langle w_f^1,w_f^2,\ldots,w_f^N \rangle$. 

The procedure for automatic cross-lingual phrase alignment consists of two stages: (1) Population of the set of possible target PP label sequences $S:\{ L_f^1, \ldots, L_f^X \}$  and (2) Scoring of the sequences in $S$ in terms of which leads to a more appropriate segmentation. Finally, the best scoring sequence is selected as the target PP sequence. %We can put a threshold here  -> future work

Possible target prosodic phrasings need to follow certain requirements in order to obtain meaningful phrase alignments. Firstly, it is necessary that $L_f$ contains the same number of unique PP labels as $L_e$ and in the same order. This is obtained by listing all possible sequences starting from $l_e^1$ and then eliminating the ones that don't follow a monotonic progression until last label $l_e^M$ is reached. Secondly, it is made sure that tokens that need to fall inside the same PP are not assigned different labels. This is especially relevant in the cases where tokens represent sub-linguistic units. 

Second stage consists of ranking each candidate PP label sequence $s = \langle l_f^1,l_f^2,\ldots,l_f^N \rangle \in S$. The alignment information residing in attention matrix holds is used to define a score as follows:
\begin{equation}
\textit{score}(s) = \prod_n^{N}{\sum_m^M{W_{\textit{masked}}^{l}(n;m)}}
\end{equation}

$W_{\textit{masked}}^{l}$ is a masked version of the attention matrix $W$ with respect to label $l$. This ensures obtaining of a higher score if the target tokens are labelled with the same label as the source tokens that they are aligned with. 

\begin{equation}
    W_{\textit{masked}}^l (i;j)=
\begin{cases}
    W(i;j) \text{if }, & l = l_e^j\\
    0,                 & \text{otherwise}
\end{cases}
\end{equation}

Finally, the best scoring PP label sequence is chosen as optimal PP label sequence. 
\begin{equation}
L_f = \argmax_{s \in S} \textit{score}(s)
\end{equation}

The ranking methodology is a way to average alignment scores in order to get a parallel phrasing structure. By defining allowed sequences first and then ranking them, complicated token alignments do not result in broken phrasings. For example, a target token that aligns with an outlier token (or doesn't align at all) gets evened out among its neighbouring tokens and thus end up in the same phrase with them.

\subsection{Synthesis timing synchronization}

\begin{figure}[t]
  \centering
  \includegraphics[width=\linewidth]{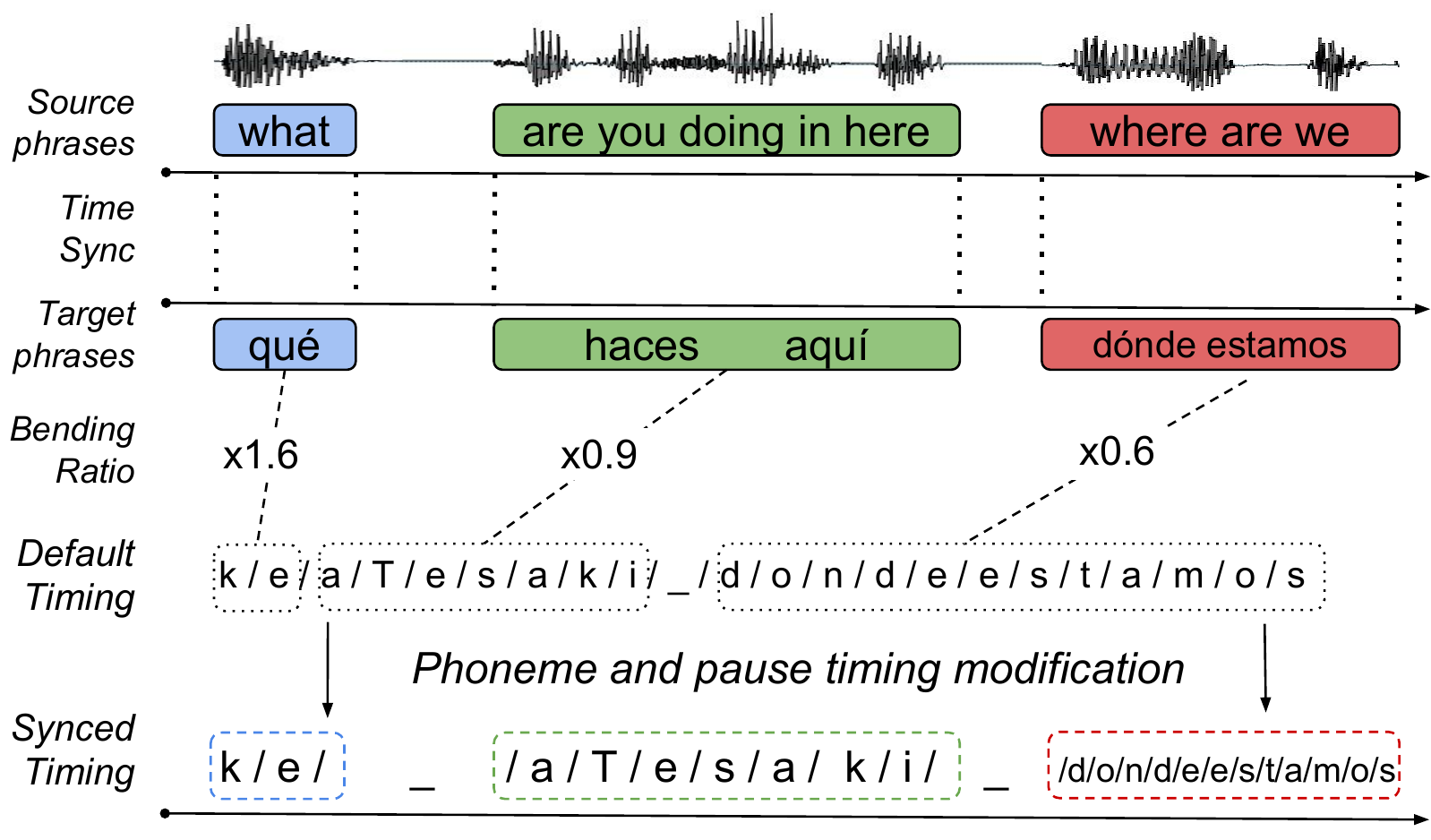}
  \caption{Durational conditioning on the target prosodic phrases with respect to source prosodic phrase and pause durations used for phoneme timing synchronization.}
  \label{fig:time_alignment}
\end{figure}

%\toni{can it happend that alignment makes jumps? Eg. English words with prep at the end?}

Synchronized audio segments are obtained by conditioning the synthesis with respect to a predefined phrasing and durational structure, as illustrated in Figure \ref{fig:time_alignment}. First step in achieving this is to define the desired durational structure of the translation by directly mapping the durational parameters of the source PPs to the target PPs that align with them. This is to ensure that voice activations overlap as much as possible so that lip movements align with the synthesized tracks. The desired durational structure is then used to condition the phoneme and pause durations that are assigned by the TTS back-end. A \textit{bending ratio} is calculated for each phrase by dividing the desired phrase duration by the duration of the same segment that TTS predicts. This ratio later determines how much the phonemes belonging to that PP need to be speed-up or slow-down. Finally, the modified phoneme durations are sent to the TTS front-end to obtain the synchronized audio segment. 

\section{Evaluation}

\subsection{Evaluation setup}

\textbf{Machine translation} models were trained initially on the WIT-TED corpus \cite{teddata} and then fine-tuned to our domain. Half of the Heroes Corpus was allocated for this purpose. As in the examples shown in previous section, we have only focused on the direction English$\rightarrow$Spanish. OpenNMT-py toolkit \cite{opennmt} was used for obtaining the models. The MT model consists of a 3-layer attentional encoder-decoder network with 512-unit bidirectional LSTMs \cite{sutskever, lstms, bahdanau} and employs beam search, coverage attention \cite{coverage} and input feeding \cite{luong}. A shared \textit{subword} vocabulary was trained for both languages using \textit{SentencePiece} byte-pair encoding (BPE) \cite{sentencepiece, bpe}. The sizes of the training corpora and the BLEU scores obtained with them on separate test sets is listed in Table \ref{table:mt}. 

\begin{table}[t]
\centering
\begin{tabular}{cccc}
\hline
\textbf{Dataset} & \textbf{\# Tokens (Train/test set)} & \textbf{BLEU (\%)} \\ \hline
\textit{WIT-TED} & 8.4M/366K & 34.04 \\
\textit{Heroes}  & 59K/62K & 24.14 \\\hline
\end{tabular}
\caption{\label{table:mt} Datasets used for training and in-domain adaptation of MT models, their training/testing set sizes and BLEU scores obtained in test sets. }
\end{table}

\noindent \textbf{Text-to-speech} system was chosen among free alternatives that would allow us to modify phoneme durations before the synthesis takes place. For this, we have opted for the \textit{mbrola} \cite{mbrola} system as a TTS back-end. Mbrola is a research-oriented TTS system that performs synthesis from pre-calculated linguistic and prosodic features without any text processing. As a front-end we used the \textit{espeak}\footnote{\url{http://espeak.sourceforge.net/}} tool to perform morpheme-to-phoneme conversion, phoneme time duration and $f0$ contour prediction.

\subsection{Analysis}

\begin{figure}[t]
  \centering
  \includegraphics[width=\linewidth]{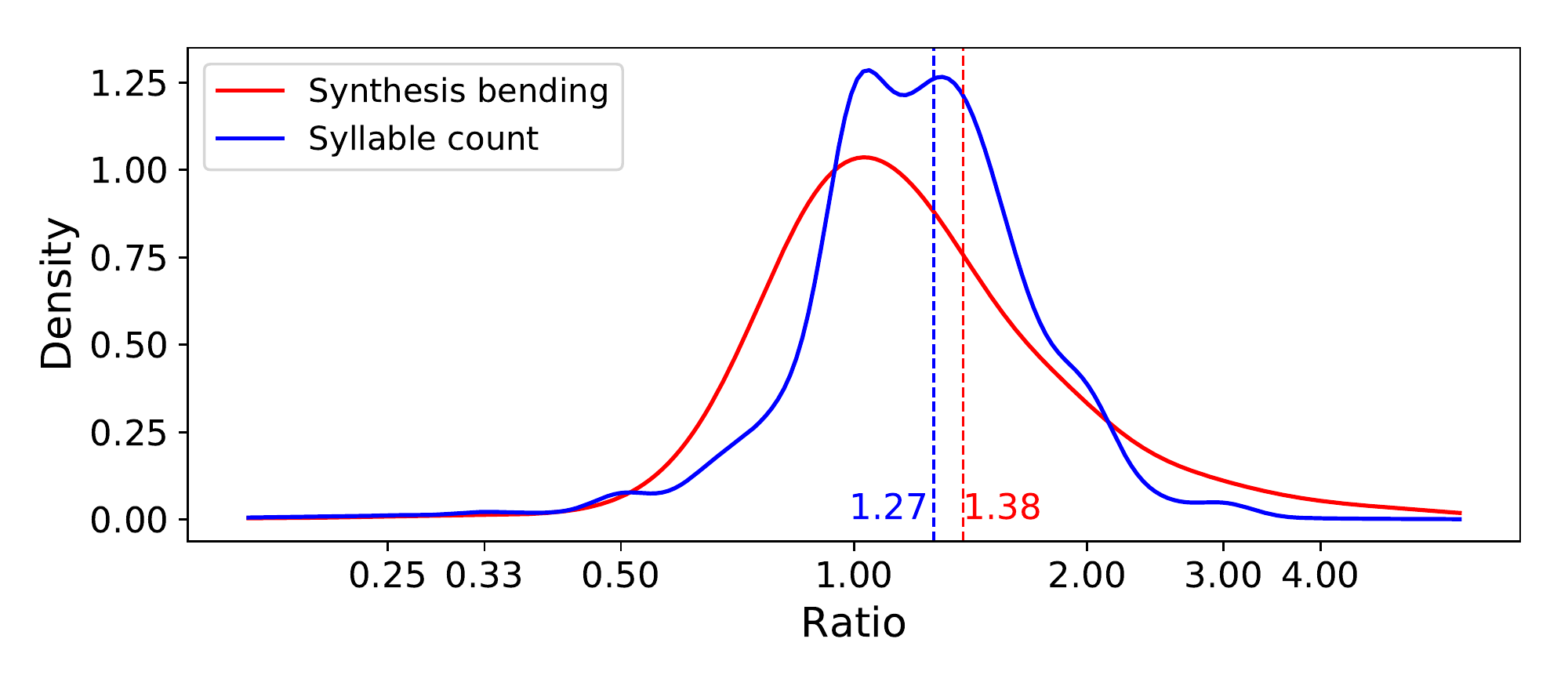}
  \caption{Speech rate comparison with respect to syllable count (blue) and synthesis time modification ratio (red) distribution between source and automatically aligned segments. Mean values are shown with dashed lines. }
  \label{fig:logratios}
\end{figure}

For evaluation, we have allocated 3490 segments of the Heroes corpus for evaluation and performed various manual and experimental analyses on them. We first wanted to see how much the aligned prosodic phrases match in terms of their length and how much of a speed-up or slow-down rate has been applied to make their durations fit during synthesis synchronization. Figure \ref{fig:logratios} shows the density graph of the speech rate ratios and bending ratios calculated for each matching prosodic phrase in the evaluation set. Speech rate ratio was calculated with respect to syllables. It can be seen that average speech rate ratio (1.27) corresponds to the original ratio measured in Heroes Corpus (1.31). It is also seen that the phrase durations predicted by the TTS front-end were more likely to be sped-up than slowed-down. 

A perception test has also been prepared to compare our methodology to a subtitle reading approach similar to \cite{czech2008, czech2010, subtts2009}. We have selected 10 samples from our evaluation set and cut out the video portions corresponding to them. Two dubbed audio tracks were prepared for each sample: One with the machine translated and synced synthesis using our methodology and another one with synthesized Spanish subtitles with their starting times dictated by their original timestamps\footnote{Samples and source code can be accessed from the project repository: \url{https://github.com/alpoktem/MachineDub}}. 18 Spanish speaking participants were asked to compare two machine dubbed versions of each sample in terms of (1) translation quality, and (2) lip-syncing precision. Also, they were asked which version they preferred for each sample. Mean opinion scores (MOS) results are shown in Table \ref{table:test}.

\begin{table}[t]
\centering
\begin{tabular}{cccc}
\hline
\textbf{System} &  \multicolumn{2}{c}{\textbf{MOS}} & \textbf{Preference} \\
 & \textbf{Translation} & \textbf{Lip-sync} &  \\ \hline
\textit{subtitle} & 4.08 & 3.44 & \%68 \\
\textit{synced}  & 2.96 & 3.58 & \%32 \\\hline
\end{tabular}
\caption{\label{table:test} Mean opinion scores (MOS) for translation and lip-syncing quality between synthesis of translated subtitle segments (\textit{subtitle}) and our methodology (\textit{synced})}.
\end{table}

\subsection{Discussion}

On manual inspection, our methodology works generally well in determining meaningful prosodic phrasing for the translations only by using the attention output. A similar speech rate ratio was obtained in average for automatically aligned phrases as the original dubbed segments. However, manual inspection on the synthesized samples show that bending ratios that are higher or lower than a certain limit lead to unnatural sounding syntheses. This calls for further modalities that could be integrated in translation to achieve similar length segments both in terms of syllable count and articulation duration. Also, the system as it is only manages to align syntheses to spoken intervals. Further lip-syncing guidelines such as alignment of open, closed phones should also be given attention to achieve further realistic synchronicity.

Another bottleneck that we observed both through manual analyses and perception test results is the effect of poor translation quality. Test participants clearly opt for professional translations even in average their lip-syncing performs more poorly. 

\vspace{-1mm}

\section{Conclusions}

In this paper, we have introduced a methodology for the synchronization of prosodic phrases in a machine dubbing scenario. Our main contribution is a simple but effective exploitation of the by-product of neural machine translation to achieve a mapping between prosodic phrases in the source sentence with tokens in the target sentence. We have further demonstrated its use in a complete machine dubbing pipeline that fuses this transferred phrasing information with the durational structure of the source sentence in order to obtain a synchronized audio translation. In average, the length-wise comparison of automatically aligned phrases outputted by our methodology was able to get close to the speech rate ratio of professionally dubbed segments. Also, qualitative evaluations suggest that it is possible to obtain acceptable or better lip-syncing quality with automatically translated and synchronized dialogue lines. Future work will involve a scoring mechanism for phoneme level alignment and also exploiting of the N-best translation results that could help obtain better speech rate ratios using alternative translations. 

\vspace{-1mm}

\section{Acknowledgements}

The second author is funded by the Spanish Ministry of Economy, Industry and Competitiveness through the \textit{Ram\'on y Cajal} program.

\bibliographystyle{IEEEtran}

\bibliography{mybib}

\end{document}